\documentclass[a4paper]{article}
\pdfoutput=1

\usepackage[inline]{enumitem}

\usepackage{IBERSPEECH2020}

\title{Domain Adaptation in Dialogue Systems using Transfer and Meta-Learning}
\name{Rui Ribeiro$^{1 2}$, Alberto Abad$^{1 2}$, José Lopes$^3$}

\address{
  $^1$INESC-ID Lisboa, Portugal\\
  $^2$Instituto Superior Técnico, Universidade de Lisboa, Portugal\\
  $^3$Heriot-Watt University, Edinburgh, United Kingdom}
\email{rui.m.ribeiro@tecnico.ulisboa.pt}

\begin{document}

\maketitle
\begin{abstract}
Current generative-based dialogue systems are data-hungry and fail to adapt to new unseen domains when only a small amount of target data is available. Additionally, in real-world applications, most domains are underrepresented, so there is a need to create a system capable of generalizing to these domains using minimal data. In this paper, we propose a method that adapts to unseen domains by combining both transfer and meta-learning (DATML). DATML improves the previous state-of-the-art dialogue model, DiKTNet, by introducing a different learning technique: meta-learning. We use Reptile, a first-order optimization-based meta-learning algorithm as our improved training method.
We evaluated our model on the MultiWOZ dataset and outperformed DiKTNet in both BLEU and Entity F1 scores when the same amount of data is available. 
 
\end{abstract}
\noindent\textbf{Index Terms}: dialogue systems, domain adaptation, transfer-learning, meta-learning

\section{Introduction}

\label{section:introduction}

With the appearance of chatbots like Siri and Alexa capable of having fluent and consistent conversations, dialogue systems have become very popular these days. 
Additionally, the emergence of deep learning techniques in natural language processing contributes to this popularity and various new models were created in order to surpass previous rule-based models. 
However, these generative-based models are data-hungry, they need large amounts of training data in order to obtain good results, they produce dull responses and fail to adapt to new unseen domains when only a few examples of data are available.
Besides, in real-world applications, most domains are underrepresented, so there is a need to create a model capable of generalizing to these domains using the minimum amount of data available.

In this paper, we study the importance of generalizing to unseen domains using minimal data and aim to design a novel model to surpass this problem.
We believe that for successful adaptation to new domains, two key features are essential for improving the overall performance of a dialogue system: better representation learning and better learning techniques.
Following this belief, we are concerned with the exploration of a method able to learn a more general dialogue representation from a large data-source and able to incorporate this information into a dialogue system. 

We follow this reasoning and introduce Domain Adaptation using Transfer and Meta-Learning (DATML), a model that combines both transfer-learning with meta-learning for the purpose of adapting to unseen domains. Our model builds upon the approach from Dialogue Knowledge Transfer Network (DiKTNet) \cite{Shalyminov2019DataEfficientGC} by enhancing its learning method while keeping the strong representation learning present in both ELMo \cite{elmo} contextual embeddings and latent representations. 
For that, we divide the training method into three training stages: 
\begin{enumerate*}
  \item A pre-training phase where some latent representations are leveraged from a domain-agnostic dataset;
  \item Source training with all data except dialogues from the target domain;
  \item Fine-tuning using few examples from the target domain.
\end{enumerate*}

We incorporate meta-learning in source training as this method proved to be promising at capturing domain-agnostic dialogue representations \cite{qian-yu-2019-domain}. However, instead of using Model-Agnostic Metal-Learning (MAML) \cite{Finn2017ModelAgnosticMF} algorithm, we use a first-order optimization-based method, Reptile \cite{reptile}, which has shown to achieve similar or even better results than MAML for low-resource NLU tasks while being more lightweight in terms of memory consumption \cite{dou-etal-2019-investigating}.

We evaluate our model on the MultiWOZ dataset \cite{Budzianowski2018MultiWOZA} and compare our approach with both Zero-Shot Dialog Generation (ZSDG) \cite{zhao-eskenazi-2018-zero} and current state-of-the-art model in few-shot dialogue generation, DiKTNet. As the code for both baselines is openly available online, we adapt and evaluate their implementations on the MultiWOZ corpus. Our model outperforms both ZSDG and DiKTNet when the same amount of data is available. Furthermore, DATML achieves superior performance with 3\% of available target data in comparison to DiKTNet with 10\%, which shows that DATML surpasses DiKTNet in terms of both performance and data efficiency.

\section{Related Work}
\label{sec:realated_work}

The reduced amount of available data has always been a problem in domain adaptation tasks. 
Methods as meta-learning \cite{Finn2017ModelAgnosticMF}, transfer-learning \cite{Long2016DeepTL,George2017DeepTL,Yao2010BoostingFT} and few-shot learning \cite{NIPS2017_6996,Satorras2017FewShotLW,Sung_2018_CVPR} were introduced to solve this problem in machine learning. 
However, there were only a few attempts to solve the problem of domain adaptation in end-to-end dialogue systems.

Perhaps, one of the first studies to pursue this direction was the work from ZSDG \cite{zhao-eskenazi-2018-zero}, where authors performed zero-shot dialogue generation using minimal data in the form of seed responses. The model is described as "zero-shot" and does not use complete dialogues, however, the model still depends on human annotated data. Although this approach seems promising, ZSDG relies on these annotations for seed responses, and in the real-world scenario, if collecting data for underrepresented domains is already difficult enough, access to annotated data becomes infeasible. 

More recent studies attempt to perform domain adaption without the need of human annotated data and adopt the methods presented above: Domain Adaptive Dialog Generation via Meta-Learning (DAML) \cite{qian-yu-2019-domain} incorporates meta-learning into the \textit{sequicity} \cite{sequicity} model to train a dialogue system able to generalize to unseen domains. This approach seems promising, yet DAML was evaluated on a synthetic dataset. DiKTNet \cite{Shalyminov2019DataEfficientGC} applies transfer learning by leveraging general latent representations from a large data-source and incorporating them into a Hierarchical Recurrent Encoder-Decoder (HRED). We will describe this model in detail in the following sections as it represents a key feature for our solution.

\section{Base Model}

As mentioned in the previous section, our base model is the work from DiKTNet \cite{Shalyminov2019DataEfficientGC}. The basic idea in DiKTNet is learning reusable latent representations from a domain-agnostic dataset and incorporate that knowledge when training using minimal data from the target domains.
DiKTNet's base model is the same from ZSDG, a HRED with an attention-based copying mechanism.

More formally, the base model's HRED \(\mathcal{F}\) is optimized according to the following loss function:

\begin{equation}\label{eq:loss_diktnet}
    \mathcal{L}_{HRED}= \log p_{\mathcal{F}^d}(\mathbf{x}_{sys}|\mathcal{F}^e(\mathbf{c}, \mathbf{x}_{usr})),
\end{equation}
where \(\mathbf{x}_{usr}\) is the user's request, \(\mathbf{x}_{sys}\) is the system's response and \(\mathbf{c}\) is the context. 

Although each domain has its specific dialogue structure, every domain still shares a general representation. Thus, the authors consider the Latent Action Encoder-Decoder (LAED) framework \cite{DBLP:journals/corr/abs-1804-08069}. 
LAED is, in essence, a Variational Auto-Encoder (VAE) representation method that allows discovering interpretable and meaningful representations of utterances into discrete latent variables. 
LAED introduces a recognition network \(\mathcal{R}\) that maps an utterance to a latent variable \(\mathbf{z}\) and a generation network \(\mathcal{G}\) that will be used to train \(\mathbf{z}\)'s representation. 
The goal is to represent the latent variable \(\mathbf{z}\) independently of the context \(\mathbf{c}\), so it can capture general dialogue semantics. 
LAED is a HRED model and the authors have introduced two versions of the model: Discrete Information Variational Auto-Encoder (DI-VAE) and Discrete Information Variational Skip-Thought (DI-VST).

DI-VAE works as a typical VAE by reconstructing the input \(\mathbf{x}\) and minimizing the error between the generated and the original data. 
The loss function that optimizes the VAE model can be described as:

\begin{equation}\label{eq:loss_divae}
\begin{split}
    \mathcal{L}_{DI-VAE}&= 
    \mathbb{E}_{q_{\mathcal{R}}(\mathbf{z}|\mathbf{x})p(\mathbf{x})}[\log p_{\mathcal{G}}(\mathbf{x} | \mathbf{z})] \\ &- KL(q(\mathbf{z}) \parallel p(\mathbf{z})),
\end{split}
\end{equation}
where \(p(\mathbf{z})\) and \(q(\mathbf{z})\) are, respectively, the prior and posterior distributions of \(\mathbf{z}\), \(KL\) is the Kullback-Leibler divergence and \(\mathbb{E}\) is the expectation.

DI-VAE model aims to capture utterance representations by reconstructing each word of the utterance. 
However, it is also possible to capture the meaning by inferring from the surrounding context, as dialogue meaning is very context-dependent.
With this, the authors propose another version, the DI-VST, which is inspired by the Skip-Thought representation \cite{Kiros2015SkipThoughtV}.  DI-VST uses the same recognition network from  DI-VAE to output the posterior distribution \(q(\mathbf{z})\), however, two generators are now used to predict both previous \(\mathbf{x}_p\) and following \(\mathbf{x}_n\) utterances. The loss function that optimizes DI-VST can now be described as:

\begin{equation}\label{eq:loss-divst}
\begin{split}
    \mathcal{L}_{DI-VST}&= \mathbb{E}_{q_{\mathcal{R}}(\mathbf{z}|\mathbf{x})p(\mathbf{x})}[\log p^n_{\mathcal{G}}(\mathbf{x}_n | \mathbf{z}) \log p^p_{\mathcal{G}}(\mathbf{x}_p | \mathbf{z})] \\ & - KL(q(\mathbf{z}) \parallel p(\mathbf{z})).
\end{split}
\end{equation}

DiKTNet learns this domain-agnostic representation from a large data-source and uses LAED models to perform this task.
DiKTNet uses the DI-VAE model to obtain a latent representation of the user's request \(\mathbf{z}_{usr}=\textrm{DI-VAE}(\mathbf{x}_{usr}\)). As for the system's response, the model also wants to predict a latent representation \(\mathbf{z}_{sys}\). In order to achieve that, DiKTNet uses the DI-VST model together with a context-aware hierarchical encoder-decoder that takes as input the user's request \(\mathbf{x}_{usr}\) and the context \(\mathbf{c}\). This encoder-decoder is different from the DI-VST for the reason that this new model, instead of predicting the previous and the following utterances, is interested in only predicting the following utterance that, in fact, is the system's response.
The authors argue that DI-VAE captures the user utterance representation and that DI-VST predicts the system's action.
When training with minimal data from the target domain, and after learning the latent representations \(\mathbf{z}_{usr}\) and \(\mathbf{z}_{sys}\), these variables are incorporated into the HRED \(\mathcal{F}\) by an updated version of the loss function from equation \ref{eq:loss_diktnet}:

\begin{equation}
\label{eq:loss_diktnet_new}
\begin{split}
    \mathcal{L}_{HRED}&=
    \mathbb{E}_{p(\mathbf{x}_{usr},c)p(\mathbf{z}_{usr},\mathbf{x}_{usr})p(\mathbf{z}_{sys} | \mathbf{x}_{usr},c)} \\
    &[\log p_{\mathcal{F}^d}(\mathbf{x}_{sys}|\{\mathcal{F}^e(\mathbf{c}, \mathbf{x}_{usr}), \mathbf{z}_{usr}, \mathbf{z}_{sys}\})],
\end{split}
\end{equation}
where \{ \} is the concatenation operator. With this, we ensure that the decoder is conditioned on the latent representations inferred in the pre-training phase and can now fine-tune in the target domain by taking into account that domain-agnostic representations. DiKTNet is also augmented with ELMo's \cite{elmo} deep contextualized representations as word embeddings.

Instead of performing joint training as in original work, we first train the model with only source domains and then fine-tune it using a few example dialogues from the target domain. Below, we present how we enhanced our base model's performance using an improved training strategy.

\section{Meta-learning}

As we referenced in section \ref{section:introduction}, better training techniques improve the overall system's performance when adapting to new unseen domains using minimal data. 
In the following sections, we present our chosen meta-learning algorithm and describe how we adapted this algorithm into our base model.

\subsection{Model-Agnostic Meta-Learning}

In section \ref{sec:realated_work}, 
we described DAML \cite{qian-yu-2019-domain} which incorporates the MAML \cite{Finn2017ModelAgnosticMF} algorithm into the \textit{sequicity} model.
This optimization-based meta-learning technique aims to learn a good initialization for the model on source domains that can be efficiently adapted to target domains using minimum fine-tuning.

More formally, in each iteration of MAML, two batches of the training corpus are sampled from a source domain \(d\): \(\mathcal{D}^d_s\) and \(\mathcal{D}^d_q\) which are named, respectively, the \textit{source} and the \textit{query} set. Instead of calculating the gradient step and updating the model, in each episode, low-resource fine-tuning is simulated: the model's parameters \(\theta\) are preserved and for each domain \(d\) in source domains, new temporary parameters are calculated according to:

\begin{equation}\label{eq:maml1}
    \theta^d= \theta - \beta \nabla_\theta\mathcal{L}(\theta, \mathcal{D}^d_s),
\end{equation}
where \(\beta\) is the inner learning rate. We could update the model's original parameters with the sum of the losses from all source domains, however, we choose to update the parameters after each domain iteration as this method performs better as presented by \cite{antoniou2018how}.

After each episode, the model's parameters are updated using the temporary ones calculated in equation \ref{eq:maml1}:

\begin{equation}\label{eq:maml2}
    \theta= \theta - \alpha \nabla_\theta\mathcal{L}(\theta^d, \mathcal{D}^d_q),
\end{equation}
where \(\alpha\) is the outer learning rate. As our model incorporates both context and knowledge-base information for each dialogue and as MAML also consumes too much memory, we instead adopt a lightweight version of the MAML algorithm that we describe below.

\subsection{Reptile}

Reptile \cite{reptile} algorithm is a first-order meta-learning algorithm where instead of sampling two source and query sets, \(k > 1\) batches are retrieved for each domain \(\mathcal{D}^d = (\mathcal{D}^d_1, ..., \mathcal{D}^d_k)\) and used to create the temporary model's parameters. The loss for the temporary model is calculated using Adam \cite{adam} optimizer according to:

\begin{equation}\label{eq:reptile1}
    \theta^d= \textrm{Adam}^k(\theta, \mathcal{D}^d, \beta) ,
\end{equation}
where \(\beta\) is the inner learning rate and \(k\) is the number of updates in \(\mathcal{D}^d\). After each episode, the model's original parameters are updated using the ones calculated in equation \ref{eq:reptile1}:

\begin{equation}\label{eq:reptile2}
    \theta= \theta + \alpha (\theta^d - \theta),
\end{equation}
where \(\alpha\) is the outer learning rate. Reptile is shown in \cite{reptile} to produce equivalent or even better updates than MAML while consuming lower memory.

\subsection{DATML}

Our final model, DATML, is an adaptation of the architecture of DiKTNet with a modified training technique, while maintaining the strong representation learning. Instead of two training stages as in original work, we split joint training into source training and fine-tuning:

\begin{enumerate}
  \item \textbf{Pre-training:} we maintain the first phase, where we learn the latent general representations for each turn using DI-VAE and DI-VST models.
  \item \textbf{Source training:} in this phase, we exclude all data from the target domain and improve the training method by employing the Reptile meta-learning algorithm.
  \item \textbf{Fine-tuning:} finally, we fine-tune the model using only a few example dialogues from the target domain.
\end{enumerate}

\section{Experiments}

In this section, we describe how we evaluated both ZSDG and DiKTNet baselines and DATML. We also analyze and suggest possible limitations of our approach.

\subsection{Datasets}

\bgroup
\def\arraystretch{1.2}
\begin{table}[h!]
\centering
\caption{Excluded domains from MetalWOZ for each target domain on MultiWOZ dataset.}
\label{table:excluded}
\begin{tabular}{|l|l|}
\hline
\begin{tabular}[c]{@{}l@{}}\textbf{MultiWOZ}\end{tabular} & \begin{tabular}[c]{@{}l@{}}\textbf{MetalWOZ}\end{tabular}                             \\ \hline
hotel                                                        & \textit{HOTEL\_RESERVE}                                                                              \\ \hline
restaurant                                                   & \textit{\begin{tabular}[c]{@{}l@{}}MAKE\_RESTAURANT\_RESERVATIONS\\ RESTAURANT\_PICKER\end{tabular}} \\ \hline
attraction                                                   & \textit{EVENT\_RESERVE}                                                                              \\ \hline
\end{tabular}
\end{table}
\egroup

The dataset used to obtain the latent actions in the pre-training phase for DiKTNet and DATML was the MetalWOZ dataset.
MetalWOZ \cite{metalwoz} is a dataset specifically constructed for the task of generalizing to unseen domains and is designed to help developing meta-learning models. 
This dataset contains about 37k task-oriented dialogues in 47 domains, such as schedules, apartment search, alarm setting, and banking. 
The data was collected in a Wizard-of-Oz fashion where a person acted like a robot/system and another acted as the user. 

Both baselines and our approach were evaluated on the three most represented domains from Multi-Domain Wizard-of-Oz dataset \cite{Budzianowski2018MultiWOZA}: hotel, restaurant and attraction, where each contains more than 1500 dialogues. MultiWOZ is a large-scale multi-domain corpus containing human-to-human conversations with rich semantic labels (dialogue acts and domain-specific slot-values) from various domains and topics, and, like MetalWOZ, was collected in a Wizard-of-Oz fashion.

\subsection{Experimental Setup}

In the pre-training stage, we choose to learn the latent representations on MetalWOZ dataset as it is a domain-agnostic corpus introduced specifically for learning general representations. In order to make the evaluation as fair as possible, we exclude all dialogues from domains on MetalWOZ that could relate with the target domain on MultiWOZ, as described in table \ref{table:excluded}.

For source training, we train DATML on MultiWOZ dataset and exclude all dialogues from the target domains, including the multi-domain dialogues that contain turns from the target domain. In the fine-tuning phase, we use low resource data that varies from \(1\%\) to \(10\%\) by following \cite{Shalyminov2019DataEfficientGC} approach. 

For both baselines and DATML, we follow \cite{zhao-eskenazi-2018-zero} and \cite{Shalyminov2019DataEfficientGC} original setting and use Adam optimizer with a learning rate of \(10^{-3}\) and Dropout \((p = 0.3)\) \cite{dropout}. All RNNs have hidden size of 512 and were trained for 50 epochs, using early stopping if the validation accuracy does not improve on half of already completed epochs.
In the pre-training phase, we train both DI-VAE and DI-VST based LAED with \(y\) size of \(10\) and \(k\) size of \(5\), where \(y\) represents the number of latent variables and \(k\) the number of possible discrete values for each variable. For Reptile, we use a \(k\) size of \(5\) and train the model for \(4000\) episodes. The inner and outer learning rates are \(10^{-3}\) and \(10^{-1}\), respectively.

For ZSDG, we followed the original author's \cite{zhao-eskenazi-2018-zero} setting and used 150 seed responses for each domain. In order to fairly compare our model with state-of-the-art DiKTNet, we choose the same domain target data for both models by setting the random seed to \(271\), with no particular reason for selecting that number.

\subsection{Metrics}

We follow the work from DiKTNet \cite{Shalyminov2019DataEfficientGC} and ZSDG \cite{zhao-eskenazi-2018-zero} and report BLEU and Entity F1 for each domain. These scores are calculated for each turn, where BLEU measures the similarity between the predicted and the reference responses and Entity F1 determines the ability of the model to retrieve correct entities from the knowledge base.

\section{Results and Discussion}

\bgroup
\def\arraystretch{1.5}
\begin{table*}[t!]
\footnotesize
\centering
\caption{Results on MultiWOZ dataset.}
\label{table:results}
\begin{tabular}{|l||c|c||c|c||c|c|}
\hline
\multicolumn{1}{|r||}{\footnotesize{\textbf{Domain}}}       & \multicolumn{2}{c||}{\textbf{hotel}} & \multicolumn{2}{c||}{\textbf{restaurant}} & \multicolumn{2}{c|}{\textbf{attraction}} \\
{\footnotesize\textbf{Model}}         & BLEU \%   & Entity F1 \%   & BLEU \%      & Entity F1 \%     & BLEU \%     & Entity F1 \%     \\ \hline \hline
ZSDG           & 5.0            & 8.0                  & 4.7                & 14.3                  & 6.0                & 16.0                   \\ \hline \hline
DiKTNet - 1\%  & 10.7            & 17.3                 & 12.4               & 17.5                   & 10.2               & 18.6                   \\ 
DiKTNet - 3\%  & 11.4             & 18.2                 & 13.4               & 26.0                   & 12.4               & 20.6                  \\
DiKTNet - 5\%  & 11.6            & 17.6                 & 16.6              & 25.7                   & 12.0               & 27.1                   \\ 
DiKTNet - 10\% & 13.1           & 16.8                 & 16.9               & 28.2                   & 12.3               & 27.4                    \\ \hline \hline
DATML - 1\%    & \textbf{10.9}   & \textbf{18.0}        & \textbf{14.1}      & \textbf{24.0}          & \textbf{11.0}      & \textbf{23.4}          \\ 
DATML - 3\%    & \textbf{13.0}   & \textbf{23.1}        & \textbf{16.7}      & \textbf{28.4}          & \textbf{14.1}      & \textbf{28.6}          \\ 
DATML - 5\%    & \textbf{14.1}   & \textbf{25.3}        & \textbf{17.8}      & \textbf{30.0}           & \textbf{15.0}      & \textbf{31.2}          \\ 
DATML - 10\%   & \textbf{14.2}   & \textbf{26.3}        & \textbf{18.3}      & \textbf{32.9}          & \textbf{15.4}      & \textbf{32.2}          \\ \hline
\end{tabular}
\end{table*}
\egroup

\begin{figure}[t]
  \centering
  \includegraphics[width=\linewidth]{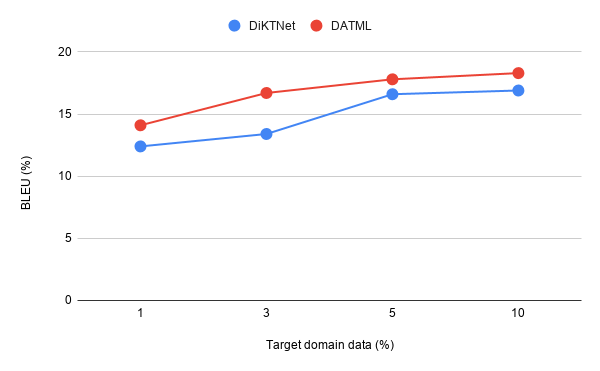}
  \caption{BLEU score for different amounts of target data in the restaurant domain.}
  \label{fig:bleu}
\end{figure}

\begin{figure}[t]
  \centering
  \includegraphics[width=\linewidth]{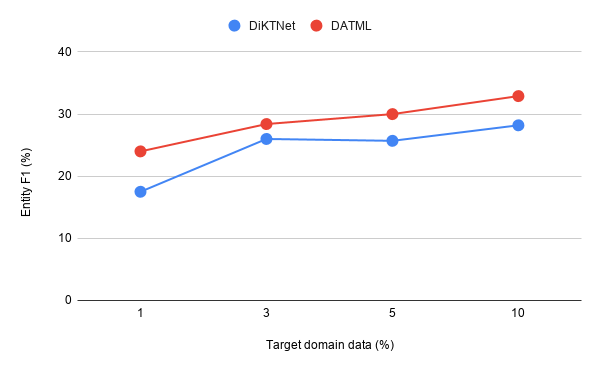}
  \caption{Entity F1 score for different amounts of target data in the restaurant domain.}
  \label{fig:f1}
\end{figure}

Table \ref{table:results} shows results on MultiWOZ dataset. As observed in bold values, DATML outperforms both baselines ZSDG and DiKTNet in all low-resource scenarios.

We investigate how the use of different amounts of target domain data has an impact in the system's performance. Table \ref{table:results} shows that our model's performance correlates with the amount of available data from the unseen domain. Figures \ref{fig:bleu} and \ref{fig:f1} reveal that correlation for the restaurant domain and compare DiKTNet and DATML in terms of data usage.
 While small improvements can be observed when only \(1\%\) of target domain data is available, for each domain DATML achieves better results with \(3\%\) of target data in all metrics in comparison to DiKTNet with \(10\%\) of available target data. This shows that DATML outperforms DiKTNet in terms of both performance and data efficiency.
   
 Table \ref{table:results} also confirms that DiKTNet and DATML outperform ZSDG while using no annotated data and thus discarding human effort in annotating dialogues. This confirms that DATML achieves state-of-the-art results in data-efficiency and that is most suitable for real-world applications, as in underrepresented domains the amount of annotated data is almost nonexistent.

 The results demonstrate that using optimization-based meta-learning improves the overall model's performance, and validate our initial idea that better learning techniques are a key feature when adapting to unseen domains using minimal data. Although our results seem promising and DATML outperforms previous state-of-the-art DiKTNet, these low scores are far from being sufficient for real-world applications, and more work is essential to surpass the problem of data scarcity in dialogue systems.

\section{Conclusions}

Domain adaptation in dialogue systems is extremely important as most domains are underrepresented. We proposed a model that improves previous state-of-the-art method by enhancing the training method. However, the evaluation results indicate that our model is far from being suited for real-world applications and show that this field requires more study. Future work includes improving the latent representations' retrieval and integration into our model. We would also like to refer that after submitting this paper we started some experiments with BERT-based \cite{BERT} embeddings which are left for future work.

\section{Acknowledgements}

This work has been partially supported by national funds through Fundação para a Ciência e a Tecnologia (FCT) with reference UIDB/50021/2020.

\bibliographystyle{IEEEtran}

\bibliography{mybib}

\begin{thebibliography}{10}
\providecommand{\url}[1]{#1}
\csname url@samestyle\endcsname
\providecommand{\newblock}{\relax}
\providecommand{\bibinfo}[2]{#2}
\providecommand{\BIBentrySTDinterwordspacing}{\spaceskip=0pt\relax}
\providecommand{\BIBentryALTinterwordstretchfactor}{4}
\providecommand{\BIBentryALTinterwordspacing}{\spaceskip=\fontdimen2\font plus
\BIBentryALTinterwordstretchfactor\fontdimen3\font minus
  \fontdimen4\font\relax}
\providecommand{\BIBforeignlanguage}[2]{{%
\expandafter\ifx\csname l@#1\endcsname\relax
\typeout{** WARNING: IEEEtran.bst: No hyphenation pattern has been}%
\typeout{** loaded for the language `#1'. Using the pattern for}%
\typeout{** the default language instead.}%
\else
\language=\csname l@#1\endcsname
\fi
#2}}
\providecommand{\BIBdecl}{\relax}
\BIBdecl

\bibitem{Shalyminov2019DataEfficientGC}
\BIBentryALTinterwordspacing
I.~Shalyminov, S.~Lee, A.~Eshghi, and O.~Lemon, ``Data-efficient goal-oriented
  conversation with dialogue knowledge transfer networks,'' in
  \emph{Proceedings of the 2019 Conference on Empirical Methods in Natural
  Language Processing and the 9th International Joint Conference on Natural
  Language Processing (EMNLP-IJCNLP)}.\hskip 1em plus 0.5em minus 0.4em\relax
  Hong Kong, China: Association for Computational Linguistics, Nov. 2019, pp.
  1741--1751. [Online]. Available:
  \url{https://www.aclweb.org/anthology/D19-1183}
\BIBentrySTDinterwordspacing

\bibitem{elmo}
\BIBentryALTinterwordspacing
M.~Peters, M.~Neumann, M.~Iyyer, M.~Gardner, C.~Clark, K.~Lee, and
  L.~Zettlemoyer, ``Deep contextualized word representations,'' in
  \emph{Proceedings of the 2018 Conference of the North {A}merican Chapter of
  the Association for Computational Linguistics: Human Language Technologies,
  Volume 1 (Long Papers)}.\hskip 1em plus 0.5em minus 0.4em\relax New Orleans,
  Louisiana: Association for Computational Linguistics, Jun. 2018, pp.
  2227--2237. [Online]. Available:
  \url{https://www.aclweb.org/anthology/N18-1202}
\BIBentrySTDinterwordspacing

\bibitem{qian-yu-2019-domain}
\BIBentryALTinterwordspacing
K.~Qian and Z.~Yu, ``Domain adaptive dialog generation via meta learning,'' in
  \emph{Proceedings of the 57th Annual Meeting of the Association for
  Computational Linguistics}.\hskip 1em plus 0.5em minus 0.4em\relax Florence,
  Italy: Association for Computational Linguistics, Jul. 2019, pp. 2639--2649.
  [Online]. Available: \url{https://www.aclweb.org/anthology/P19-1253}
\BIBentrySTDinterwordspacing

\bibitem{Finn2017ModelAgnosticMF}
\BIBentryALTinterwordspacing
C.~Finn, P.~Abbeel, and S.~Levine, ``Model-agnostic meta-learning for fast
  adaptation of deep networks,'' in \emph{Proceedings of the 34th International
  Conference on Machine Learning}, ser. Proceedings of Machine Learning
  Research, D.~Precup and Y.~W. Teh, Eds., vol.~70.\hskip 1em plus 0.5em minus
  0.4em\relax International Convention Centre, Sydney, Australia: PMLR, 06--11
  Aug 2017, pp. 1126--1135. [Online]. Available:
  \url{http://proceedings.mlr.press/v70/finn17a.html}
\BIBentrySTDinterwordspacing

\bibitem{reptile}
A.~{Nichol}, J.~{Achiam}, and J.~{Schulman}, ``{On First-Order Meta-Learning
  Algorithms},'' \emph{arXiv e-prints}, p. arXiv:1803.02999, Mar. 2018.

\bibitem{dou-etal-2019-investigating}
\BIBentryALTinterwordspacing
Z.-Y. Dou, K.~Yu, and A.~Anastasopoulos, ``Investigating meta-learning
  algorithms for low-resource natural language understanding tasks,'' in
  \emph{Proceedings of the 2019 Conference on Empirical Methods in Natural
  Language Processing and the 9th International Joint Conference on Natural
  Language Processing (EMNLP-IJCNLP)}.\hskip 1em plus 0.5em minus 0.4em\relax
  Hong Kong, China: Association for Computational Linguistics, Nov. 2019, pp.
  1192--1197. [Online]. Available:
  \url{https://www.aclweb.org/anthology/D19-1112}
\BIBentrySTDinterwordspacing

\bibitem{Budzianowski2018MultiWOZA}
\BIBentryALTinterwordspacing
P.~Budzianowski, T.-H. Wen, B.-H. Tseng, I.~Casanueva, S.~Ultes, O.~Ramadan,
  and M.~Ga{\v{s}}i{\'c}, ``{M}ulti{WOZ} - a large-scale multi-domain
  {W}izard-of-{O}z dataset for task-oriented dialogue modelling,'' in
  \emph{Proceedings of the 2018 Conference on Empirical Methods in Natural
  Language Processing}.\hskip 1em plus 0.5em minus 0.4em\relax Brussels,
  Belgium: Association for Computational Linguistics, Oct.-Nov. 2018, pp.
  5016--5026. [Online]. Available:
  \url{https://www.aclweb.org/anthology/D18-1547}
\BIBentrySTDinterwordspacing

\bibitem{zhao-eskenazi-2018-zero}
\BIBentryALTinterwordspacing
T.~Zhao and M.~Eskenazi, ``Zero-shot dialog generation with cross-domain latent
  actions,'' in \emph{Proceedings of the 19th Annual {SIG}dial Meeting on
  Discourse and Dialogue}.\hskip 1em plus 0.5em minus 0.4em\relax Melbourne,
  Australia: Association for Computational Linguistics, Jul. 2018, pp. 1--10.
  [Online]. Available: \url{https://www.aclweb.org/anthology/W18-5001}
\BIBentrySTDinterwordspacing

\bibitem{Long2016DeepTL}
\BIBentryALTinterwordspacing
M.~Long, H.~Zhu, J.~Wang, and M.~I. Jordan, ``Deep transfer learning with joint
  adaptation networks,'' in \emph{Proceedings of the 34th International
  Conference on Machine Learning}, ser. Proceedings of Machine Learning
  Research, D.~Precup and Y.~W. Teh, Eds., vol.~70.\hskip 1em plus 0.5em minus
  0.4em\relax International Convention Centre, Sydney, Australia: PMLR, 06--11
  Aug 2017, pp. 2208--2217. [Online]. Available:
  \url{http://proceedings.mlr.press/v70/long17a.html}
\BIBentrySTDinterwordspacing

\bibitem{George2017DeepTL}
D.~George, H.~Shen, and E.~Huerta, ``Deep transfer learning: A new deep
  learning glitch classification method for advanced ligo,'' \emph{arXiv
  preprint arXiv:1706.07446}, 2017.

\bibitem{Yao2010BoostingFT}
Y.~Yao and G.~Doretto, ``Boosting for transfer learning with multiple
  sources,'' \emph{2010 IEEE Computer Society Conference on Computer Vision and
  Pattern Recognition}, pp. 1855--1862, 2010.

\bibitem{NIPS2017_6996}
\BIBentryALTinterwordspacing
J.~Snell, K.~Swersky, and R.~Zemel, ``Prototypical networks for few-shot
  learning,'' in \emph{Advances in Neural Information Processing Systems 30},
  I.~Guyon, U.~V. Luxburg, S.~Bengio, H.~Wallach, R.~Fergus, S.~Vishwanathan,
  and R.~Garnett, Eds.\hskip 1em plus 0.5em minus 0.4em\relax Curran
  Associates, Inc., 2017, pp. 4077--4087. [Online]. Available:
  \url{http://papers.nips.cc/paper/6996-prototypical-networks-for-few-shot-learning.pdf}
\BIBentrySTDinterwordspacing

\bibitem{Satorras2017FewShotLW}
V.~Garcia and J.~Bruna, ``Few-shot learning with graph neural networks,''
  \emph{arXiv preprint arXiv:1711.04043}, 2017.

\bibitem{Sung_2018_CVPR}
F.~Sung, Y.~Yang, L.~Zhang, T.~Xiang, P.~H. Torr, and T.~M. Hospedales,
  ``Learning to compare: Relation network for few-shot learning,'' in
  \emph{Proceedings of the IEEE conference on computer vision and pattern
  recognition}, 2018, pp. 1199--1208.

\bibitem{sequicity}
\BIBentryALTinterwordspacing
W.~Lei, X.~Jin, M.-Y. Kan, Z.~Ren, X.~He, and D.~Yin, ``{S}equicity:
  Simplifying task-oriented dialogue systems with single sequence-to-sequence
  architectures,'' in \emph{Proceedings of the 56th Annual Meeting of the
  Association for Computational Linguistics (Volume 1: Long Papers)}.\hskip 1em
  plus 0.5em minus 0.4em\relax Melbourne, Australia: Association for
  Computational Linguistics, Jul. 2018, pp. 1437--1447. [Online]. Available:
  \url{https://www.aclweb.org/anthology/P18-1133}
\BIBentrySTDinterwordspacing

\bibitem{DBLP:journals/corr/abs-1804-08069}
\BIBentryALTinterwordspacing
T.~Zhao, K.~Lee, and M.~Eskenazi, ``Unsupervised discrete sentence
  representation learning for interpretable neural dialog generation,'' in
  \emph{Proceedings of the 56th Annual Meeting of the Association for
  Computational Linguistics (Volume 1: Long Papers)}.\hskip 1em plus 0.5em
  minus 0.4em\relax Melbourne, Australia: Association for Computational
  Linguistics, Jul. 2018, pp. 1098--1107. [Online]. Available:
  \url{https://www.aclweb.org/anthology/P18-1101}
\BIBentrySTDinterwordspacing

\bibitem{Kiros2015SkipThoughtV}
R.~Kiros, Y.~Zhu, R.~R. Salakhutdinov, R.~Zemel, R.~Urtasun, A.~Torralba, and
  S.~Fidler, ``Skip-thought vectors,'' in \emph{Advances in Neural Information
  Processing Systems}, C.~Cortes, N.~Lawrence, D.~Lee, M.~Sugiyama, and
  R.~Garnett, Eds., vol.~28.\hskip 1em plus 0.5em minus 0.4em\relax Curran
  Associates, Inc., 2015, pp. 3294--3302.

\bibitem{antoniou2018how}
\BIBentryALTinterwordspacing
A.~Antoniou, H.~Edwards, and A.~J. Storkey, ``How to train your {MAML},''
  \emph{CoRR}, vol. abs/1810.09502, 2018. [Online]. Available:
  \url{http://arxiv.org/abs/1810.09502}
\BIBentrySTDinterwordspacing

\bibitem{adam}
D.~P. Kingma and J.~Ba, ``Adam: A method for stochastic optimization,''
  \emph{arXiv preprint arXiv:1412.6980}, 2014.

\bibitem{metalwoz}
``A dataset of multi-domain dialogs for the fast adaptation of conversation
  models,'' \url{https://www.microsoft.com/en-us/research/project/metalwoz/}.

\bibitem{dropout}
\BIBentryALTinterwordspacing
N.~Srivastava, G.~Hinton, A.~Krizhevsky, I.~Sutskever, and R.~Salakhutdinov,
  ``Dropout: A simple way to prevent neural networks from overfitting,''
  \emph{Journal of Machine Learning Research}, vol.~15, no.~56, pp. 1929--1958,
  2014. [Online]. Available:
  \url{http://jmlr.org/papers/v15/srivastava14a.html}
\BIBentrySTDinterwordspacing

\bibitem{BERT}
\BIBentryALTinterwordspacing
J.~Devlin, M.~Chang, K.~Lee, and K.~Toutanova, ``{BERT:} pre-training of deep
  bidirectional transformers for language understanding,'' \emph{CoRR}, vol.
  abs/1810.04805, 2018. [Online]. Available:
  \url{http://arxiv.org/abs/1810.04805}
\BIBentrySTDinterwordspacing

\end{thebibliography}

\end{document}